%% file: main.tex
\documentclass[Afour,sageh,times]{sagej}

\usepackage{moreverb,url}
\usepackage{amssymb}

\newcommand\red[1]{\textcolor{red}{#1}}

\usepackage{siunitx}

\usepackage{pifont} 
\newcommand{\xmark}{\ding{55}}%

\usepackage{makecell}
\usepackage{multirow}

\newcommand\thline{\specialrule{.1em}{0.5em}{0.5em}}
\newcommand\pfirst{$1^{\text{st}}$}
\newcommand\psecond{$2^{\text{nd}}$}

\usepackage{svg}

\usepackage{gensymb} 
\usepackage{xfrac} 
\usepackage{graphicx} 

\usepackage[shortcuts, acronym]{glossaries}
\newacronym{slam}{SLAM}{Simultaneous Localization and Mapping}
\newacronym{lo}{LO}{Lidar Odometry}
\newacronym{lio}{LIO}{Lidar-Inertial Odometry}
\newacronym{gnss}{GNSS}{Global Navigation Satellite System}
\newacronym{rtkgnss}{RTK GNSS}{Real-Time Kinematics GNSS}
\newacronym{imu}{IMU}{Inertial Measurement Unit}
\newacronym{fov}{FoV}{Field of View}
\newacronym{pps}{PPS}{Pulse Per Second}
\newacronym{ins}{INS}{Inertial Navigation System}
\newacronym{fog}{FOG}{Fiber Optic Gyroscope}
\newacronym{rlg}{RLG}{Ring Laser Gyroscope}
\newacronym{icp}{ICP}{Iterative Closest Points}
\newacronym{bev}{BEV}{Bird's Eye View}
\newacronym{wsv}{WSV}{Whitespace Separated Value}
\newacronym{mems}{MEMS}{Micro-Electro-Mechanical Systems}
\newacronym{mocap}{MOCAP}{Motion Capture}

\newacronym{ape}{APE}{Absolute Pose Error}
\newacronym{rpe}{RPE}{Relative Pose Error}

\makeatother

\usepackage[colorlinks,bookmarksopen,bookmarksnumbered,citecolor=red,urlcolor=red]{hyperref}
\usepackage{subfig}

\newcommand\BibTeX{{\rmfamily B\kern-.05em \textsc{i\kern-.025em b}\kern-.08em
T\kern-.1667em\lower.7ex\hbox{E}\kern-.125emX}}

\def\volumeyear{2026}

\begin{document}

\runninghead{Kurda et al}

\title{Odyssey: An Automotive Lidar-Inertial Odometry Dataset with GNSS-denied situations}

\author{Aaron Kurda\affilnum{1}, Simon Steuernagel\affilnum{1}, Lukas Jung\affilnum{2} and Marcus Baum\affilnum{1}}

\affiliation{\affilnum{1}University of Göttingen, Göttingen, Germany\\
\affilnum{2}iMAR Navigation, St. Ingbert, Germany}

\corrauth{Aaron Kurda, Institute of Computer Science
University of Göttigen,
Göttingen, Germany}

\email{aaron.kurda@cs.uni-goettingen.de}

\begin{abstract}
The development and evaluation of \ac{lio} and \ac{slam} systems requires a precise ground truth. The \ac{gnss} is often used as a foundation for this, but its signals can be unreliable in obstructed environments due to multi-path effects or loss-of-signal. While existing datasets compensate for sporadic \ac{gnss} loss by incorporating \ac{imu} measurements, the commonly used systems do not permit prolonged study of \ac{gnss}-denied environments due to accumulated drift. Therefore, the diversity of such datasets is limited. To close this gap, we present Odyssey, an automotive \ac{lio} dataset featuring: (1) a ground truth derived from a navigation-grade \ac{rlg}-based RTK/INS, offering bias stability one to four orders of magnitude better than existing automotive datasets; (2) a comprehensive collection of 36 sequences across diverse environments, enabling robust and comprehensive evaluation and (3) prolonged \ac{gnss}-denied environments, including tunnels and, previously unseen in the context of automotive benchmarks, indoor parking garages. Here, our \ac{rlg}-based system enables accurate evaluation in scenarios where commonly employed systems would drift excessively.
Besides providing data for \ac{lio}, Odyssey also supports place recognition tasks through threefold trajectory repetition and integration of external mapping data via precise geodetic coordinates.
All data, dataloader and supplementary material are available online at \url{https://odyssey.uni-goettingen.de/}.

\end{abstract}


\maketitle

\glsresetall

\section{INTRODUCTION}
\begin{figure*}[t]
    \centering  \includegraphics[width=0.98\linewidth]{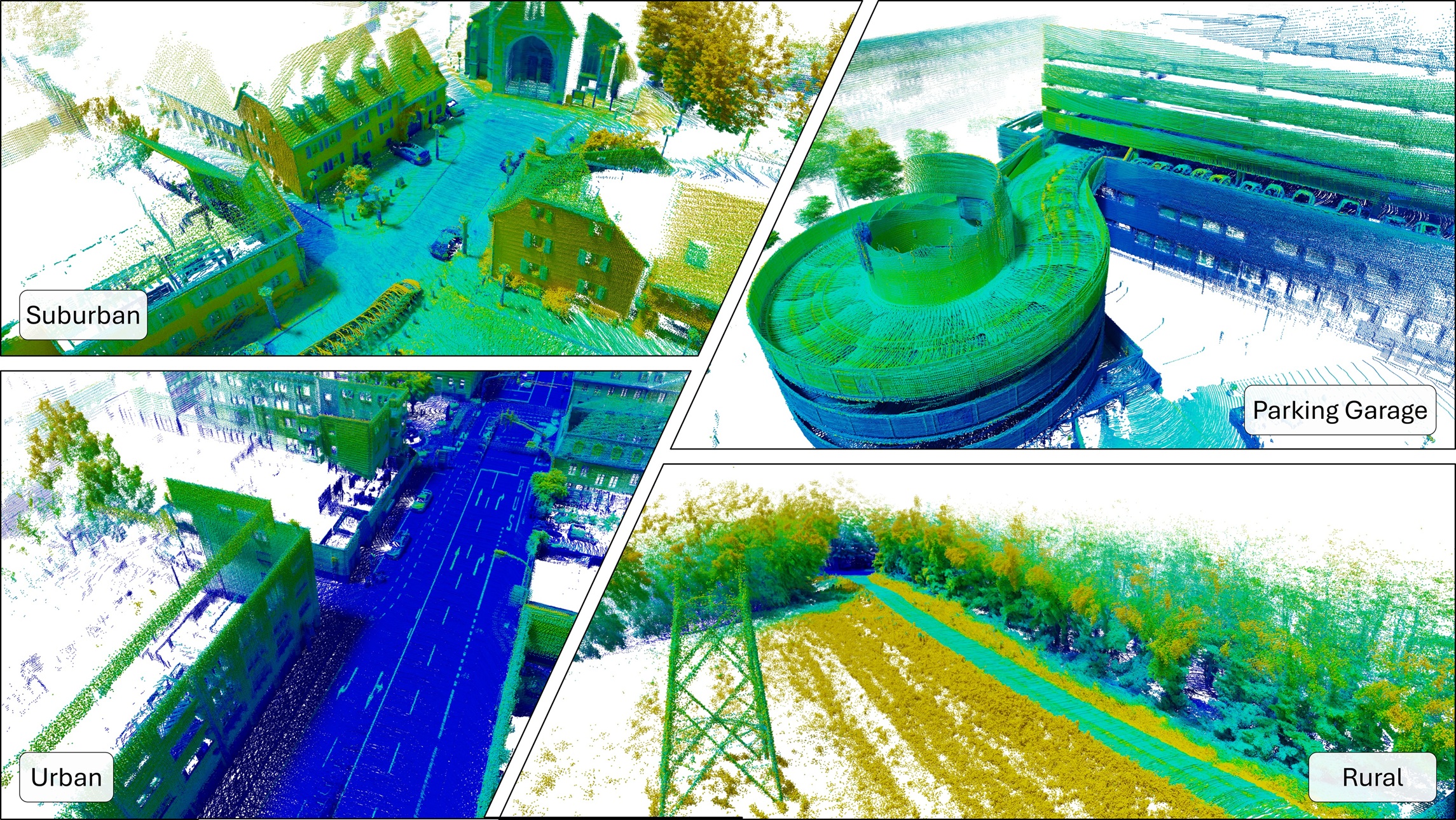}
    \caption{
    Example subset of the variety of environments included in the Odyssey dataset: 
    Standard urban, suburban and rural environments as well as long-term \acrshort{gnss}-denied situation such as indoor parking garages and tunnels.
    Images were created by accumulating lidar scans based on the ground truth data.}    \label{fig:titlepicture}
\end{figure*}
Accurate and reliable localization is a key task for autonomous mobile systems. For outdoor applications such as autonomous driving,  this is typically achieved via~\ac{gnss}, which is based on relative position measurements to satellites. This, however, requires the robust reception of weak radio signals from suitable satellite constellations, which is not feasible in certain environments such as urban canyons or tunnels. Beyond environmental obstructions, these signals are also vulnerable to security threats and may even be actively disrupted by malicious actors. For these reasons, it is vital for autonomous vehicles to be able to continue accurate positioning in situations with (partial) \ac{gnss} loss.

A common approach in \ac{gnss}-denied situations is to solely rely on an \ac{imu}, which allows the calculation of the ego-pose through the integration of linear acceleration and angular velocity measurements. This is known as strapdown inertial navigation~\citep{budiyono2012principles}. 
Another common choice is to employ lidar data to estimate the ego-motion of a mobile platform. This is called lidar odometry~\citep{cadena2016past, lee2024lidar} or \ac{slam}~\citep{zhang20243d} depending on the focus of the work.
Recent works have explored the fusion of lidar odometry with inertial navigation, giving rise to the problem of \ac{lio}~\citep{shan2020liosam,malladi2025robust}. While more robust, \ac{lio} methods suffer from the same problem as lidar odometry and inertial navigation: the accumulation of errors.

Two other directions that try to alleviate or entirely circumvent this problem have arisen in recent years. Map-assisted lidar odometry~\citep{frosi2023osm,kurda2025reducing,banon2022robust,banon2024geo} tries to reduce the drift of odometry methods through the integration of prior map data.
While effective in suitable environments, these methods require a strong initial guess for successful registration against the prior map. Place-recognition methods~\citep{yin2025general} the other hand try to circumvent this problem entirely. Such methods aim to estimate the pose of a vehicle by comparing a single lidar scan against a database of previously recorded scans, allowing it to recognize previously seen places without the need for a strong initial guess~\citep{yan2019global,cho2022openstreetmap}. Compared to the map-assisted odometry methods, however, these approaches often require feature-rich, dense point cloud maps of the environment for robust and accurate localization~\citep{dube2020segmap}.

Regardless of the chosen localization approach, all methods rely on suitable data for their development and evaluation. Beyond the lidar and inertial data for running a \ac{lio} algorithm, an accurate ground truth is required for assessing performance. In many outdoor scenarios such ground truth can be obtained using a \ac{gnss}-augmentation system~\citep{budiyono2012principles}, e.g., \ac{rtkgnss}. This, in combination with additional sensors such as \acp{imu}, can provide an accurate pose estimate independent of length and duration. Such devices can be referred to as~RTK/INS. Receiving \ac{gnss} signals, however, requires a clear view towards the sky and an absence of reflecting surfaces. This is not only a problem in tunnels or parking garages, but already (comparably) small buildings or short underpasses can cause a sporadic drop in accuracy. 
While alternative landmark-based localization systems, such as roadside units, can be employed in smaller, controlled environments, equipping larger areas with such infrastructure quickly becomes infeasible. 
If no measurements to known landmarks can be received, the estimated position will slowly start to accumulate the measurement errors of the \acp{imu}, resulting in a drifting trajectory~\citep{budiyono2012principles}. While this drift cannot be eliminated (without measurements to known landmarks), it can be reduced through the use of more accurate sensors.
The gyroscope is of special interest, as its quality influences both drifting behavior in GNSS-denied situations and orientation accuracy in general. Due to the strong influence of orientation on commonly used error metrics such as \ac{rpe} and KITTI metric~\citep{geiger2013vision}, accurately estimating orientation enables both longer GNSS-denied navigation and more accurate evaluation overall.

Existing automotive datasets rely on \ac{mems} or \ac{fog}-based gyroscopes for the ground truth generation (Tab.~\ref{tab:datasets}). While sufficient for GNSS-enabled scenarios, already short GNSS outages can cause a significant amount of drift which prohibits accurate evaluation in prolonged GNSS-denied environments.

\begin{figure*}[t]
    \centering
    \subfloat[\centering \texttt{ParkingGarage}]{\includegraphics[height=0.365\textwidth]{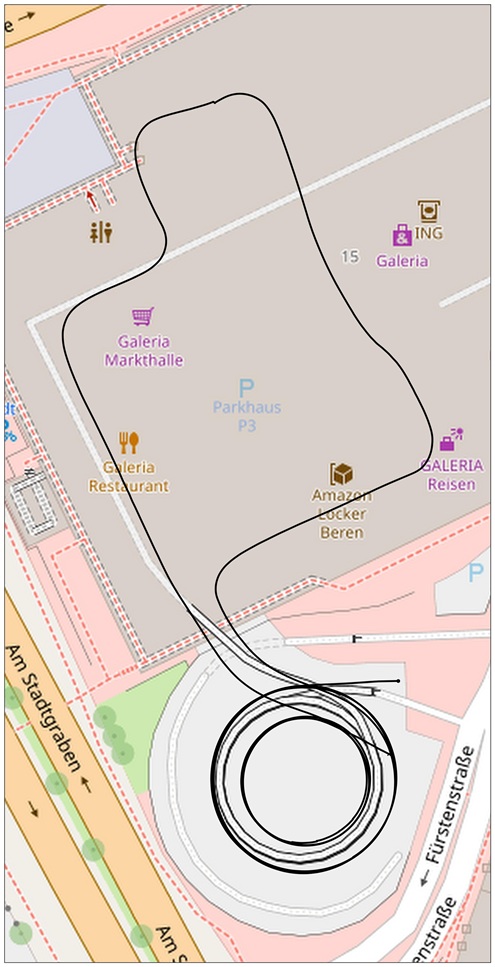}\label{fig:trajectories:parkhaus}}
    \subfloat[\centering \texttt{Marketplace}]{\includegraphics[height=0.365\textwidth]{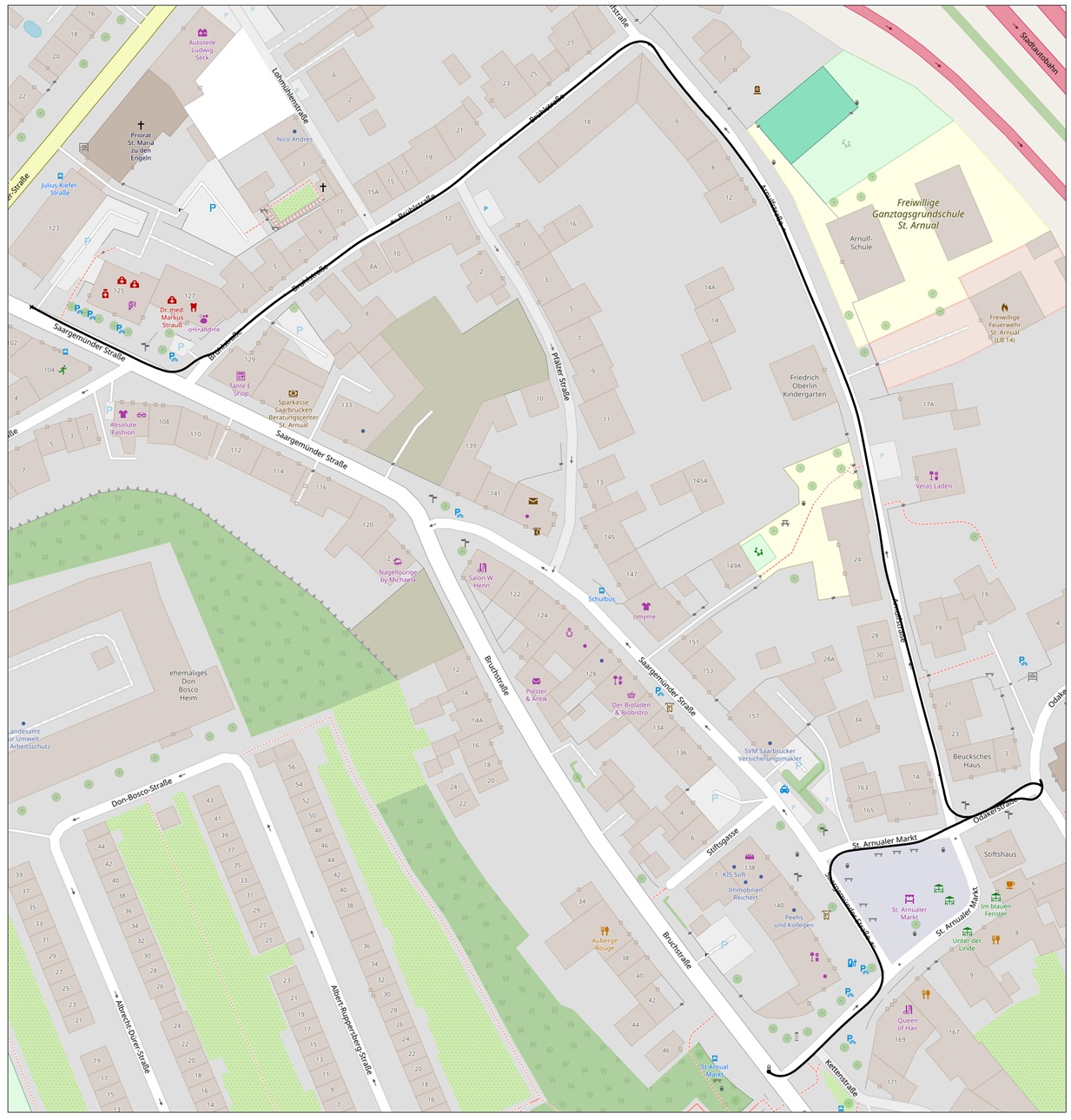}\label{fig:trajectories:marktplatz}}
    \subfloat[\centering \texttt{Suburb}]{\includegraphics[height=0.365\textwidth]{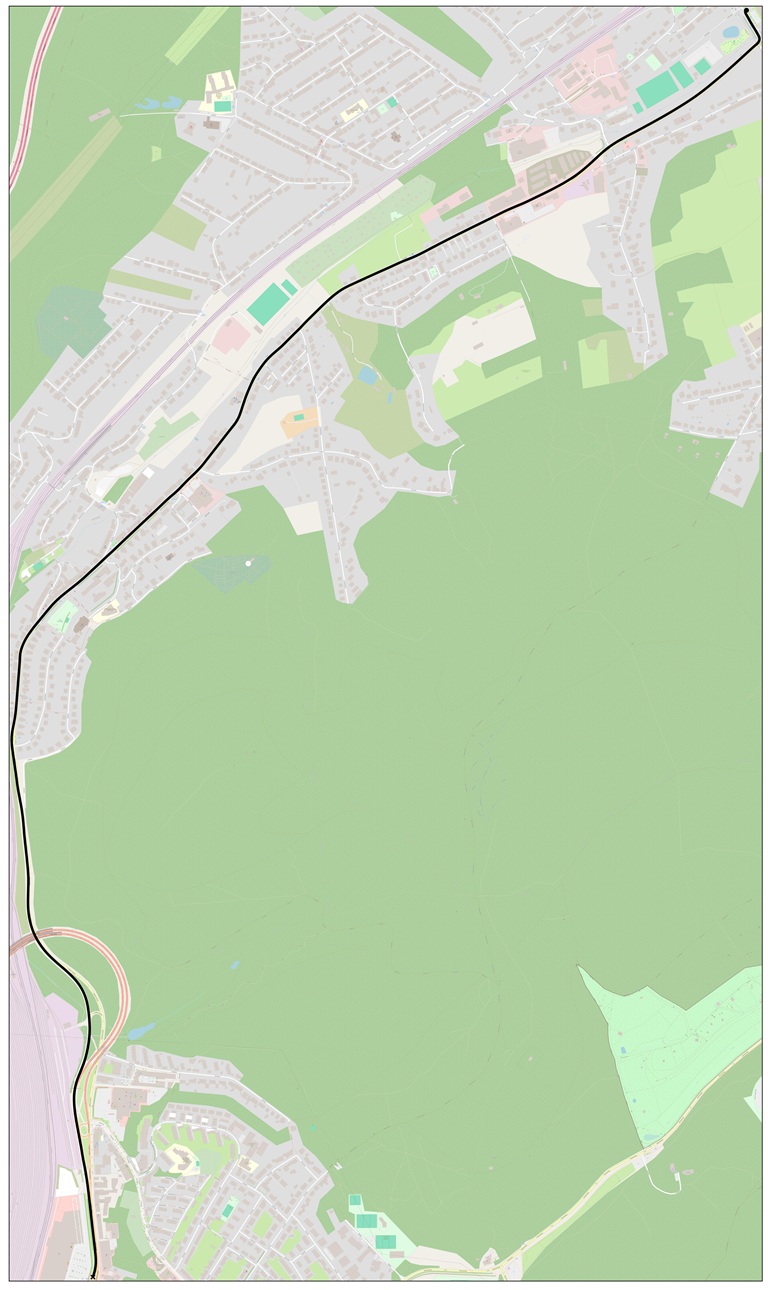}\label{fig:trajectories:vorstadt}}
    \subfloat[\centering \texttt{CountryRoad}]{\includegraphics[height=0.365\textwidth]{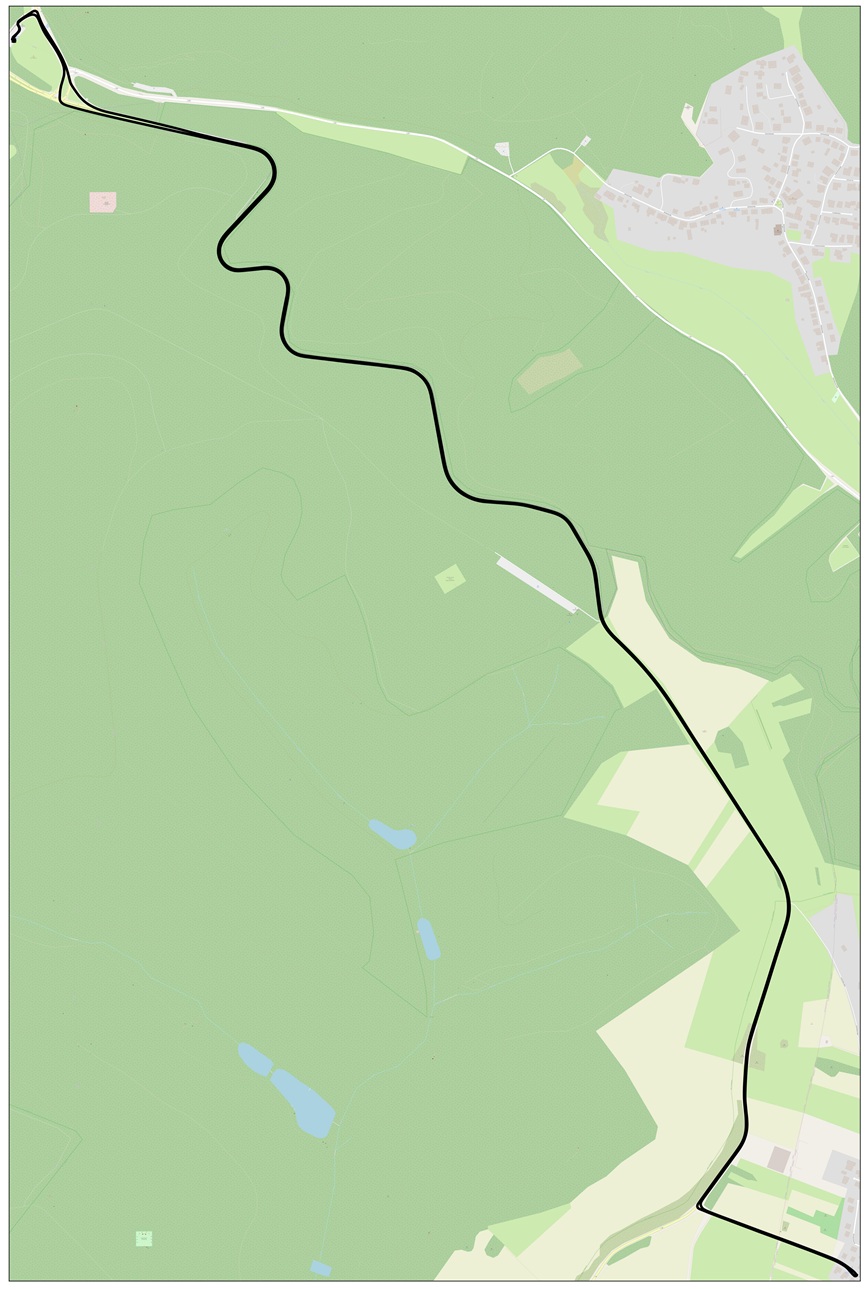}\label{fig:trajectories:landstrasse}}
    \caption{A subset of the trajectories overlayed with a rendering from OpenStreetMap, showing our dataset in urban, suburban and rural regions as well as in an indoor parking garage.}
    \label{fig:trajectories}
\end{figure*}

\subsection{Contribution}

In this article, we present Odyssey, an automotive dataset tailored towards \ac{lio} and related localization tasks.
The main contributions of Odyssey are:
\begin{itemize}
    \item An accurate ground truth derived from a navigation-grade \ac{rlg}-based RTK/INS,
    \item a large collection of 36 sequences, enabling comprehensive evaluation of methods in a diverse set of environments,
    \item and prolonged GNSS-denied environments, such as indoor parking garages, previously unseen in the context of automotive \ac{lio} datasets.
\end{itemize}
The accuracy of the ground truth data has been validated by verifying the alignment with official cadastre maps.
Besides the ground truth, Odyssey contains lidar data from an Ouster OS1 rev.7 and inertial measurements from a secondary, automotive-grade \ac{mems}-based \ac{imu}. By recording the same trajectory multiple times, our dataset also allows a clean separation between training and testing for data-driven approaches and the development of place recognition tasks that require large amounts of overlap. 
Data and supplementary material is hosted on our website
\begin{center}
\url{https://odyssey.uni-goettingen.de/}
\end{center}
including a Python dataloader, summary statistics and videos to aid in the development of new methods. Furthermore, this article provides baseline results for several state-of-the-art lidar, lidar-inertial and map-assisted lidar odometry methods.

\subsection{Structure}
The remainder of this article is structured as follows: First, related work, including existing datasets for similar purposes, is discussed in Section~\ref{sec:related_work}. 
Section~\ref{sec:dataset} gives a detailed description of our dataset, including short textual descriptions and summary statistics of the individual sequences (Section~\ref{sec:trajectories})
and a detailed description of our hardware, time synchronization and calibration processes (Section~\ref{sec:sensorsuite}). This is followed by a variety of experiments validating the accuracy of our reference system and the derivation of our ground truth (Section~\ref{sec:validation}) and
a description of the used data formats (Section~\ref{sec:dataformat}). 
Lastly, in Section~\ref{sec:evaluation}, we provide baseline results for some publicly available lidar odometry, \ac{lio} as well as map-assisted lidar odometry methods.

\section{RELATED WORK}\label{sec:related_work}

\input{tables/overview_trajectories}

One of the most established datasets is the KITTI Vision Benchmark Suite~\citep{geiger2013vision}, which also supports many other tasks such as visual odometry, optical flow or tracking tasks besides lidar odometry. Its lidar odometry (sub)-dataset comprises $22$ sequences with a strong urban focus. However, due to the absence of a secondary \ac{imu}, this dataset is not suitable for \ac{lio}.
ComplexUrban~\citep{jeong2019complex} and MulRan~\citep{kim2020mulran} are two more recent datasets. Opposed to KITTI, these include angular velocities and linear accelerations from a separate \ac{imu}. For ground truth generation, both datasets rely on a combination of \ac{rtkgnss}, \ac{fog} and manual post-processing using loop closure detection of lidar scans. MulRan was originally designed for place-recognition and therefore features a threefold repetition of only four unique trajectories, resulting in $12$ individual sequences. The main characteristic of the ComplexUrban dataset is its sensor arrangement, with four different rotating lidar sensors configured to maximum environmental coverage.
A related dataset to ComplexUrban and MulRan is the Hong Kong UrbanNav~\citep{hsu2023hongkong} dataset, which provides data of four different scenarios with a focus on urban-canyons. Similar to our dataset, it also features a long tunnel with few geometric features. It contains data from three spinning lidar scanners, a stereo camera and a secondary \ac{mems}-based \ac{imu}. The ground truth is again derived from a fusion from \ac{rtkgnss} and \ac{fog}-based \ac{imu} measurements.
ComplexUrban and UrbanNav provide simultaneous data from multiple lidar sensors that are arranged to deliver a comprehensive view of the surrounding environment. In contrast, the LIBRE dataset~\citep{carballo2020libre} is designed to enable a systematic comparison between different lidar scanners. It uses a single mounting position, where the lidar sensor is exchanged between individual data acquisition runs. The dataset includes $10$ distinct lidar scanners. Furthermore, \ac{imu} and \ac{gnss} measurements are available, although the dataset does not provide explicit ground-truth trajectories. HeLiPR~\citep{jung2024helipr} is a dataset tailored towards studying place recognition using different types of lidar scanners. Alongside its three original trajectories, it integrates seamlessly with MulRan by supplying new recordings of environments previously released as part of the MulRan dataset. Besides the data from four different spinning as well as solid-state lidars, the dataset contains a ground truth derived from a \ac{mems}-based \ac{ins} and additional inertial data. Similar to this, the EU long-term dataset~\citep{yan2020eu} offers data from $11$ different lidar, radar and camera sensors with the goal of providing heterogeneous data for studying long-term autonomy for intelligent vehicles including adverse weather conditions among others. The ground truth is again derived from \ac{gnss} and \ac{imu} measurements.
Adverse weather conditions have also been covered by the CADC~\citep{pitropov2021canadian} and the Boreas~\citep{burnett2023boreas} datasets. The CADC dataset contains $75$ very short sequences of length between $50$ and $100$ lidar scans making it only partially usable for studying the drifting behavior of odometry methods. The Boreas dataset on the other hand contains $45$ long trajectories of the same environment. While the evaluation of \ac{lio} methods is certainly possible, the single trajectory limits the needed diversity for a robust evaluation. While \acp{rlg} have been used to evaluate methods in past works~\citep{georgy2012vehicle}, there currently does not exist a publicly available dataset with ground truths derived from such a system.

In contrast to the car-mounted setup used for our dataset, a variety of datasets using other platforms such as mobile robots, handheld devices or backpack-mounted systems have been proposed.
The TIERS dataset~\citep{qingqing2022multi} is recorded from a custom-built sensor platform that can either be mounted on a robot for motorized operation or manually pushed, depending on the environment. The platform carried 5 spinning and solid-state lidar scanners as well as a lidar camera. No \ac{imu} or \ac{gnss} receivers were installed and the ground truth was instead obtained using \ac{mocap} and \ac{slam} systems. The dataset contains both outdoor and indoor scenarios, with the indoor scenarios limited to office areas and hallways.
The Newer College~\citep{ramezani2020newer} and, more recently, the Oxford Spires~\citep{tao2025spires} datasets were captured using a handheld device. They cover indoor and outdoor environments, with ground truth generated by registering lidar scans against a prior map obtained from a static secondary lidar.
A similar dataset was proposed in~\citep{jin2025largescale}, containing lidar and inertial data collected using a custom backpack-mounted platform in both structured and unstructured environments. Unlike the Newer College and Oxford Spires datasets, its ground truth is derived from a fusion of GNSS measurements and lidar-based mapping. Synthetic datasets generated using high-fidelity simulation, such as CarlaScenes~\citep{kloukiniotis2022carlascenes}, have also been proposed. For the purposes of this work, however, we focus exclusively on real-world data and therefore do not discuss such datasets further.

\section{THE ODYSSEY DATASET}\label{sec:dataset}
The Odyssey dataset is divided into 12 trajectories, with each trajectory repeated three times, resulting in a total of 36 sequences. Rather than focusing on an individual environment as often done in modern datasets~\citep{jeong2019complex, kim2020mulran}, the deliberate decision was made to collect data from a diverse set of environments, as shown in Fig.~\ref{fig:titlepicture} and Fig.~\ref{fig:trajectories}. These range from urban canyons in densely inhabited cities, over suburbs with wide streets and detached houses, to rural regions with few to no artificial buildings. This diversity ensures our dataset captures the wide range of scenarios a vehicle may encounter in everyday traffic.
In addition to such well-known scenarios, our dataset also contains rough, bumpy terrain and stop-and-go traffic for testing inertial odometry under sudden shocks and repeated acceleration and decelerations as well as challenging environments for lidar odometry systems, such as wide open fields, tight forest roads with thick vegetation, tunnels and parking garages with spiraling ramps. In total, our dataset comprises approximately $\SI{163}{\kilo\meter}$ of drive length and $\SI{4}{\hour}$ of drive duration. 

\input{tables/overview_datasets_and_ins}

\begin{figure*}
    \centering
    \newcommand\vehiclewidth{0.8}
    \subfloat[\centering Side View]{\includegraphics[width=\vehiclewidth\columnwidth]{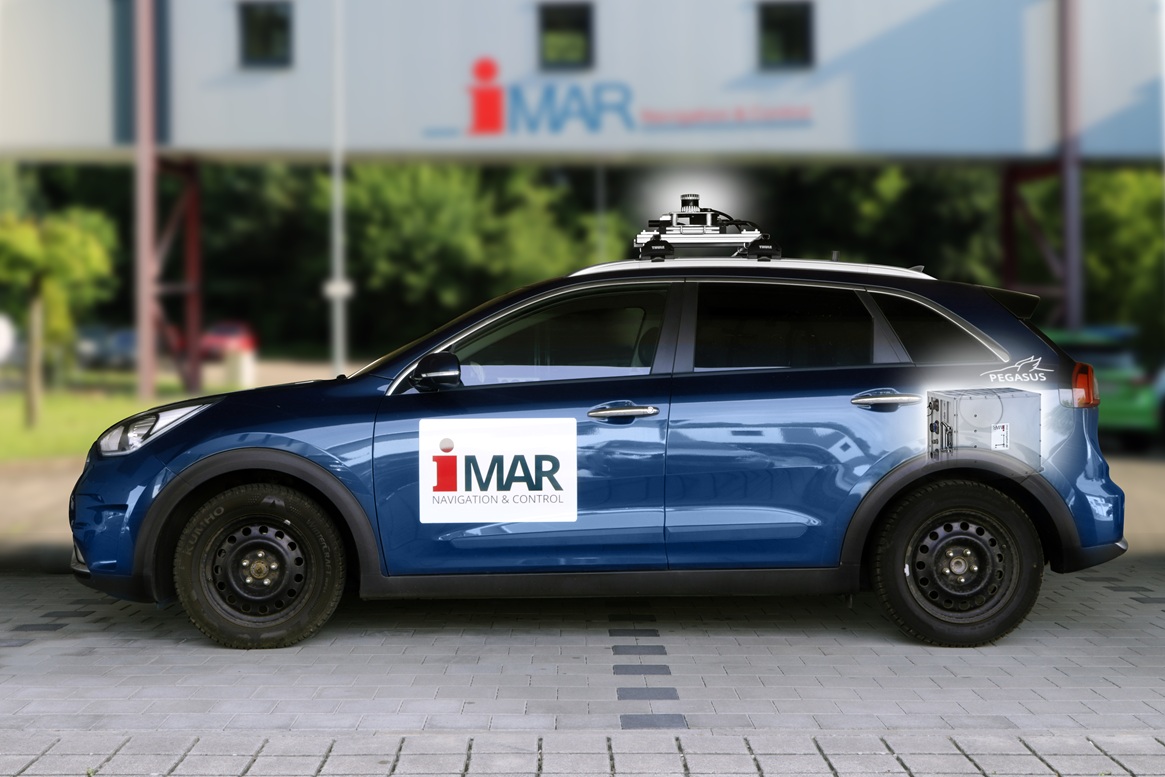}\label{fig:recording_vehicle:side}}
    \hspace{1cm}
    \subfloat[\centering Trunk View]{\includegraphics[width=\vehiclewidth\columnwidth]{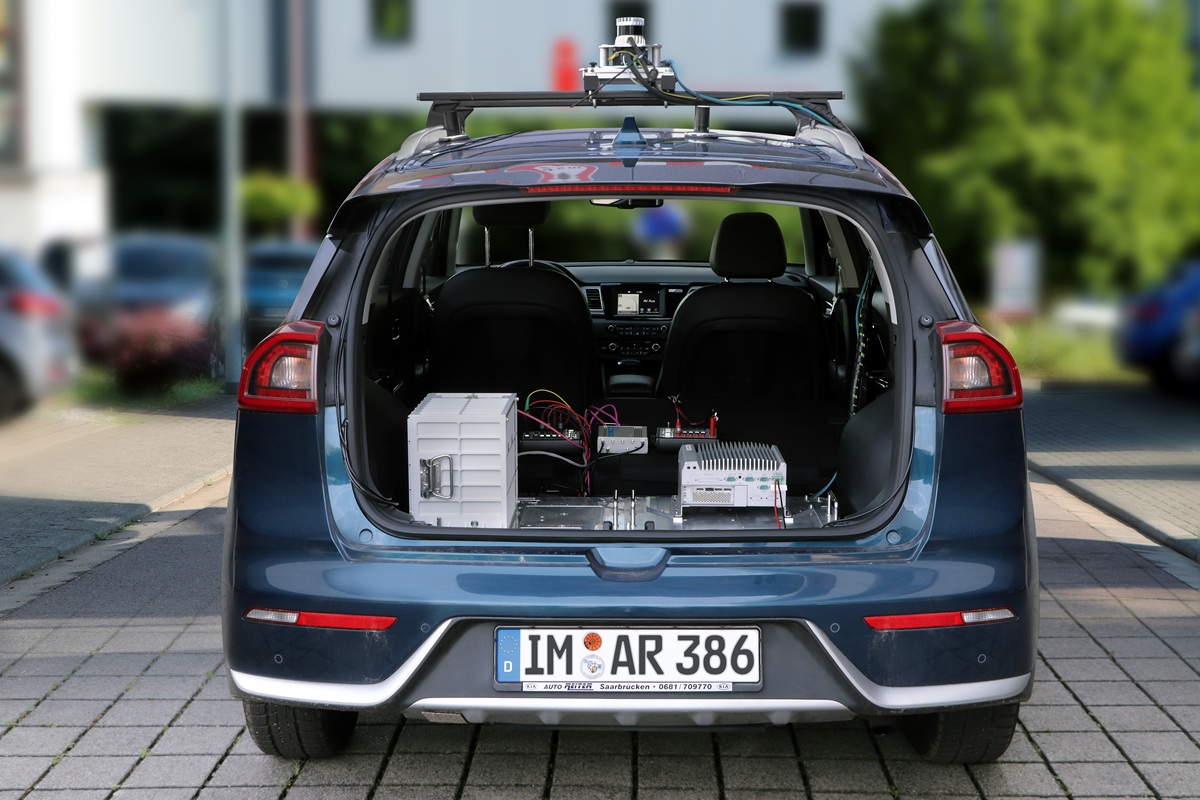}\label{fig:recording_vehicle:trunk}}\hfill
    \caption{
    Vehicle and sensor setup used for recording the data, with the $360^\circ$ lidar and estimate \acrshort{ins} on top and the reference \acrshort{ins} as well as the data collection system in the trunk. (Photos by iMAR Navigation GmbH)
    }
    \label{fig:recording_vehicle}
\end{figure*}

\subsection{TRAJECTORY DESCRIPTION}\label{sec:trajectories}
For easy identification, we use a similar nomenclature to the MulRan~\citep{kim2020mulran} and HeLiPR~\citep{jung2024helipr} datasets. For every distinct place/trajectory, we define a descriptive shorthand that reflects its content. An overview of all our trajectories including a short textual description as well as summary statistics can be found in the following enumeration as well as Tab.~\ref{tab:trajectories}. For an even more detailed analysis, we provide rendered videos of the lidar data for all sequences on our homepage.
\begin{itemize}
    \item \texttt{Beltway}: A drive starting in the city and entering a beltway, featuring high speeds and a moderate amount of other traffic participants.
    \item \texttt{CountryRoad}: A drive that features a stretch of road through a forested region with a high embankment, followed by a wide open field with few trees and bushes. 
    \item \texttt{ForestRoad}: A short round trip in and around a forest. It features thick vegetation, rough terrain and almost no artificial structures. 
    \item \texttt{Highway}: A drive over a highway that starts in a suburban area. The drive features a moderate amount of traffic and some geometry deprived environments such as bridges.
    \item \texttt{HighwayTunnel}: A long drive over a highway. It contains a long circular tunnel with no geometry for robust point cloud registration.
    \item \texttt{InnerCity}: An urban drive with many other traffic participants through the inner city of Saarbrücken.
    \item \texttt{Marketplace}: A suburban drive featuring a section of bumpy cobblestone road. 
    \item \texttt{ParkingGarage}: An indoor parking garage with a spiraling entry.
    \item \texttt{Suburb}: A long trajectory that alternates between suburban and highway environments.
    \item \texttt{Theater}: An urban drive with comparably wide roads and a moderate amount of traffic.
    \item \texttt{Tunnel}: A short city tunnel without any geometry for longitudinal localization.
    \item \texttt{UndergroundCarPark}: A underground car park, similar to \texttt{ParkingGarage}.
\end{itemize}
\subsection{SENSOR SUITE}\label{sec:sensorsuite}

The sensor suite for the Odyssey dataset contains
\begin{itemize}
    \item an iPRENA-M-II~\citep{datasheet-m2}, an \ac{rlg}-based navigation-grade RTK/INS,
    \item an iNAT M300-TLE-LN1~\citep{datasheet-m300}, an automotive-grade \ac{mems}-based RTK/INS,
    \item and an Ouster OS1 rev.7~\citep{datasheet-ouster}, a 128-layer lidar.
\end{itemize}

Together, the iNAT M300-TLE-LN1 and Ouster OS1 form the \textit{productive} subsystem that provides all data necessary for running \ac{lio} algorithms, i.e., lidar point clouds and \ac{imu} measurements. 
Data for evaluation, i.e., the ground truth, is provided by an entirely separate \textit{reference} system, the iPRENA-M-II RTK/INS.

\subsubsection{Productive System}
The productive system contains the 128-layer Ouster OS1 rev.7~\citep{datasheet-ouster} lidar and a iNAT M300-TLE-LN1~\citep{datasheet-m300} RTK/INS. Both devices are mechanically fixed to each other and rigidly mounted on the roof of the vehicle (Fig.~\ref{fig:recording_vehicle:trunk}). 
The iNAT M300-TLE-LN1 produces inertial measurements with a frequency of $\SI{300}{\hertz}$. Its output frame is already calibrated to the coordinate frame of the Ouster OS1 rev.7.
This extrinsic calibration was determined using an in-house calibration process and is guaranteed to be accurate up to $<\SI{0.1}{\milli\meter}$ for translation and $<\SI{0.06}{\degree}$ for rotation.  
For precise time keeping, the lidar is phase-synchronized to the \ac{pps}-signal of the iNAT M300-TLE-LN1 RTK/INS. It is operated at $\SI{10}{\hertz}$, yielding $128 \cdot 2048$ points per revolution.

\begin{figure*}[t]
    \centering
    \includegraphics[width=0.98\textwidth]{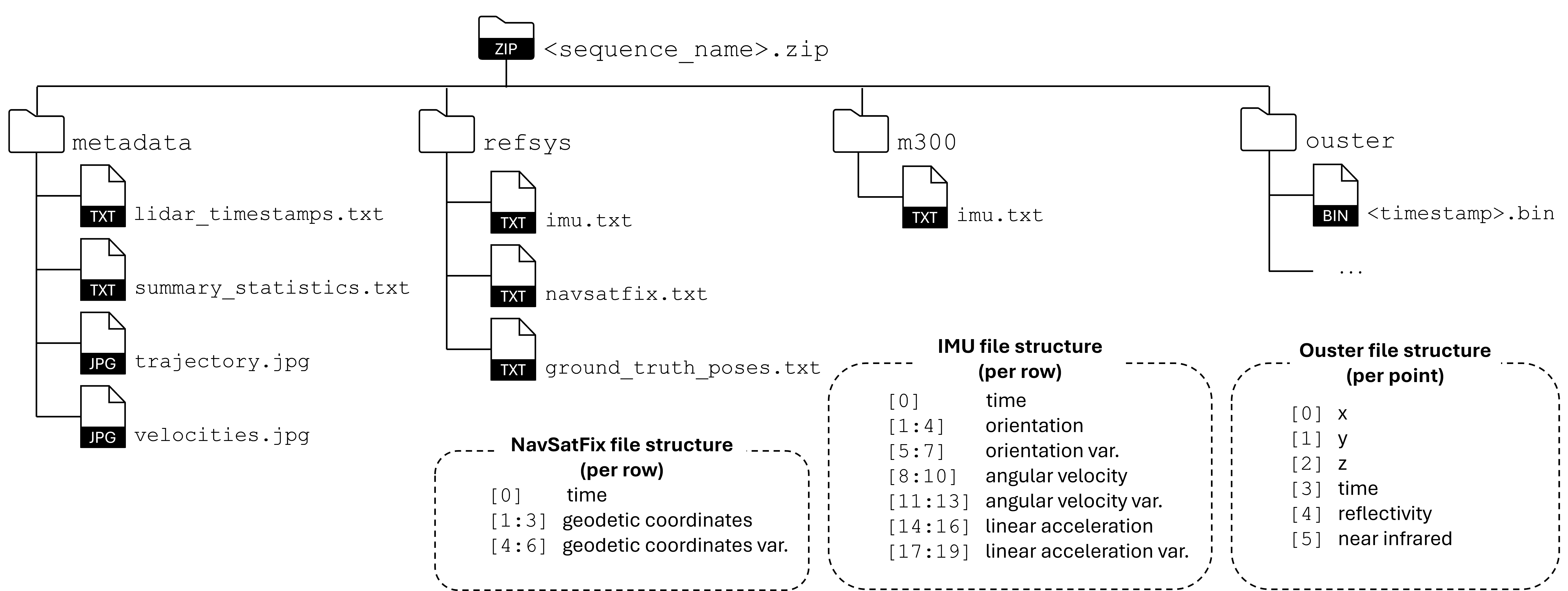}
    \caption{
    Folder and file structure of the Odyssey dataset. All data is divided by sensor. The folder \texttt{metadata} contains useful data and visualization of the sequence and the other folders contain the actual sensor data.
    }
    \label{fig:folder_structure}
\end{figure*}

\subsubsection{Reference System} \label{sec:reference_system}
The iPRENA-M-II~\citep{datasheet-m2} is a navigation-grade RTK/INS featuring a \ac{rlg} for precise orientation estimation. 
It produces pose and inertial estimates at a frequency of $\SI{200}{\hertz}$ through internal fusion of \ac{rtkgnss} and \ac{imu} measurements.
The main benefit of a \ac{rlg} as its ability to accurately estimate its orientation over prolonged periods of time, which is one of the biggest sources of error in \ac{gnss}-denied environments.
Here, both the bias instability as well as the sensor noise of our \ac{rlg} as at least one order of magnitude lower compared to the best gyroscopes used in other datasets (Table~\ref{tab:datasets}). This increased accuracy enables us to provide more accurate pose estimates in \ac{gnss}-denied environments as well as more accurate orientational estimates in general.
Physically, the system is mounted in the trunk of the vehicle as shown in Fig.~\ref{fig:recording_vehicle:trunk}, but the system's data output frame is already transformed to the lidar's origin. 
This extrinsic calibration was determined based on the mechanical design specifications of the car, followed by a hand-eye calibration process between the reference system and the iNAT M300-TLE-LN1. The resulting calibration was already applied to the data; no manual calibration has to be applied for evaluation.

\subsubsection{Intrinsic calibration}
Both the iPRENA-M-II and iNAT M300-TLE-LN1 are intrinsically calibrated using calibrated and geodetically referenced precision turntables. During this process, the inertial sensor error model in parametrized by exposing the system to precisely defined rotational rates and orientations. The calibration procedure identifies and compensates intrinsic sensor error characteristics such as bias, scale factor deviations, axis misalignments, non-orthogonalities and temperature-dependent effects. The resulting parameters are incorporated into the sensor model and used by the navigation algorithms to maximize accuracy and long-term stability.

\subsubsection{Time Synchronization}
All sensors are time-synchronized through a multi-layer approach. Both INS systems (iPRENA-M-II and iNAT M300-TLE-LN1) are equipped with high-precision internal oscillators that are synchronized through GNSS signals. 
Over a duration of $4.5$~h in a controlled GNSS-denied environment, the clock drift between both systems was measured to be $\SI{25}{\micro\second}$, corresponding to a maximum of \SI{0.41}{\micro\second} deviation during the longest GNSS-denied segment of our dataset. This effectively makes the time synchronization error between the two systems negligible in the context of \ac{lio}. The lidar itself is phase-synchronized to the \ac{pps}-signal of the iNAT M300-TLE-LN1, thus ensuring hardware-level synchronization.

\subsection{REFERENCE SETUP VALIDATION AND GROUND TRUTH GENERATION} \label{sec:validation}
To validate the accuracy and the reported uncertainties of the reference system, we conducted three complementary experiments designed to assess inter-run consistency, absolute accuracy in urban environments, and inertial drift in \ac{gnss}-denied segments. These experiments demonstrate the reported uncertainty bounds to align with the observed errors across varying environmental conditions. Building upon this validated foundation, the trajectories were further refined using loop-closure constraints to minimize residual drift and ensure intra-sequence consistency.

\subsubsection{Inter-run Consistency} \label{sec:experiment1}
To evaluate the trajectory consistency across multiple repetitions, we employed a systematic sampling strategy.
Every 50th lidar scan from the first repetition was selected, and the spatially closest neighbors from the second and third repetitions were identified. These paired scans were aligned using ICP and the resulting transformations were analyzed as a measure of inter-run consistency. Scan pairs with distances $>\SI{3}{\meter}$ were discarded to avoid potentially unstable registrations. Across 987 registrations, the mean absolute translational displacement was measured to be $[0.076,0.075,0.066]$ meter in the NWU (North-West-Up) directions and the mean absolute rotations around the XYZ-axis was measured to be $[0.0011, 0.0010, 0.0004]$ radians. $96.35\%$ of all samples fall within the $2\sigma$ uncertainty bounds of the reference system, thereby validating the reported variances of our reference system. As the measured values inherently comprise both actual measurement error as well as ICP registration noise, these reported values represent an upper bound on the true trajectory inconsistency. All in all, these results show that our reference system produces both accurate pose measurements and uncertainties that are consistent between individual repetitions.

\subsubsection{Absolute Accuracy in Urban Environments}
To assess absolute accuracy, we utilized official building outlines from cadastre data as an external reference\footnote{\url{https://geoportal.saarland.de/}}. As this data lacks elevation information, validation was restricted to horizontal (XY) components and the yaw angle. We first transformed all sequences and the building map into a common coordinate system and then registered every 50th lidar scan against the building map. To filter out regions with no or too few buildings, we only accepted ICP registrations with at least 3000 correspondences under 3 meters at initialization. The resulting set contained 574 transformations, spread over all urban regions of the dataset. 
Averaging the translational magnitude of all transformations results in an estimate for the global translational uncertainty of $\sigma_{total} = \SI{0.27}{\meter}$. 
It is important to note that $\sigma_{total} = \sqrt{\sigma^2_{ref} + \sigma^2_{map} + \sigma^2_{icp}}$, i.e., the sum of reference system, map, and ICP registration uncertainties, represents a very conservative upper bound for the true uncertainty $\sigma^2_{ref}$ of our reference system, as both the building map as well as the registration process itself introduce additional (unquantifiable) errors.

\subsubsection{Drift Analysis in GNSS-Denied Segments} \label{sec:drift}
To quantify the inertial drift specifically within GNSS-denied environments, we analyzed the trajectory segments where satellite reception was completely lost. We focused on the \texttt{ParkingGarage} and \texttt{UndergroundCarPark} sequences, as these are the longest indoor scenarios without GNSS availability. 
For each repetition of these trajectories, we extracted lidar scans at the entry/exit point of the GNSS-denied segment.
As both the entry and exit point are very close to each other, we used ICP to determine the relative transformation between them. This transformation serves as a measure of the accumulated error during the denied interval, effectively closing the loop between entering and exiting the environment.
The absolute translation errors were averaged across all repetition of both trajectories, resulting in mean absolute translation of
$[0.17,0.22,0.32]$ meter in the NWU (North-West-Up) directions, an increase of $\SI{0.30}{\meter}$ meter over the baseline errors reported in Section~\ref{sec:experiment1}\footnote{Note that this endpoint error is expected to be higher than the \acrshort{ape} as it is not averaged across the entire trajectory.}. 
Dividing this by the average distance of $\SI{450}{\meter}$ results in an average drift of $0.067\%$. We find the measured drift to be an order of magnitude lower than the \ac{rpe} (over sequences of length $\SI{450}{\meter}$) reported by multiple state-of-the-art lidar odometry and \ac{lio} methods, which renders the measured errors negligible for evaluating relative pose performance in these scenarios.
Generally, all individual measurements fall within the reported $2\sigma$ uncertainty bounds of the reference RTK/INS, confirming the specified uncertainties even during extended periods without GNSS updates.

\subsubsection{Ground Truth Generation via Loop Closures} \label{sec:ground_truth}
Building upon the validated performance of the reference system, we applied post-processing optimization to minimize residual drift in trajectories containing loops. Similar to method established in MulRan~\citep{kim2020mulran} and ComplexUrban~\citep{jeong2019complex}, we manually identified loop-closure opportunities and derived corresponding constraints using ICP. These constraints were fused with the validated RTK/INS measurements from the reference system within the factor graph optimization framework GTSAM~\citep{gtsam}, yielding an improved intra-sequence consistency and reducing the loop-closure errors of Section~\ref{sec:drift}. For trajectories without loop closures, the ground truth relies solely on the navigation-grade \ac{ins} performance. The post-processed ground truth is available in \texttt{refsys/ground\_truth\_poses.txt} file.

\subsubsection{Reference System Limitations}
We observed that during prolonged stationary phases, specifically in the \texttt{ParkingGarage1} and \texttt{ParkingGarage2} sequences, the reference system performed an internal bias refinement. This process can result in discrete discontinuities in the output trajectory. We have identified these situations and provide the start and end timestamps of these discontinuities in the \texttt{metadata} folder.

\subsection{DATA FORMAT}
\label{sec:dataformat}

The dataset is divided into 36 different sequences, labeled with a name corresponding to its trajectory, followed by a number indicating its repetition. The data of one sequence is sorted by sensor, as depicted in Fig.~\ref{fig:folder_structure}, with folder \texttt{refsys} containing ground truth data from the reference system, \texttt{m300} containing \ac{imu} measurements from the productive system and \texttt{ouster} containing the point cloud from the lidar sensor. We additionally provide a folder \texttt{metadata} that contains useful data, statistics and visualizations for each sequence. The remainder of this section gives a detailed description of the data format necessary for reading the dataset. For convenience, we also provide a dataloader written in Python on our website.

\subsubsection{IMU data format} \hfill\\
All \ac{imu} data from both the productive and the reference system are saved as a single, both human- and computer-readable \ac{wsv} file.
A single line corresponds to one \ac{imu} measurement, containing in order: the Unix timestamp \texttt{t} in nanoseconds since the 01.01.1970, the (filtered) orientation \texttt{[ori\_x, ori\_y, ori\_z, ori\_w]} specified as a unit-quaternion with trailing scalar, the uncertainties in the orientation specified through the variances \texttt{[cov\_ori\_00, cov\_ori\_11, cov\_ori\_22]}, the angular velocities \texttt{[angvel\_x, angvel\_y, angvel\_z]} and the corresponding variances \texttt{[cov\_angvel\_00, cov\_angvel\_11, cov\_angvel\_22]} and lastly
the linear accelerations \texttt{[linacc\_x, linacc\_y, linacc\_z]} and the corresponding variances \texttt{[cov\_linacc\_00, cov\_linacc\_11, cov\_linacc\_22]}. Both the angular rates as well as the linear accelerations are provided raw, i.e., without any additional filtering. Only the in-factory calibrated sensor errors are compensated to minimize potential adverse effects arising from filter cascading in real-world \ac{lio} systems. The
resulting quantities (orientations and positions), however,
are filtered during the recording process to account for the estimated sensor errors.

\subsubsection{NavSatFix data format} \hfill\\
The position data from the reference system is saved as a single, both human- and computer-readable \ac{wsv} file.
A single line corresponds to one \ac{gnss} measurement, containing the 6 fields
\texttt{[t, lat, lon, alt, cov\_00, cov\_11, cov\_22]} with
\texttt{t} being the Unix time in nanoseconds,
\texttt{lat}, \texttt{lon} and \texttt{alt} being the geodetic latitude and longitude in degrees, \texttt{alt} being the altitude above sea level in meters and \texttt{[cov\_00, cov\_11, cov\_22]} being the variances of \texttt{lat}, \texttt{lon} and \texttt{alt} respectively. The position \texttt{[lat, lon, alt]} is an already filtered quantity from the fusion of the \ac{gnss}-signals and \ac{imu} measurements. The reference coordinate system for all geodetic coordinates is ETRS89/ETRF2014.

\subsubsection{Lidar data format} \hfill\\
Every full revolution of the lidar is saved as a binary file \texttt{ouster/<timestamp>.bin} with the filename indicating the Unix time in nanoseconds. Every lidar scan contains exactly $128 \times 2048$ lidar points sorted by laser (top to bottom) and time of acquisition. A single point is defined through a densely packed array of six 32-bit floating-point numbers 
\texttt{[x, y, z, t, reflectivity, near\_ir]} with \texttt{x}, \texttt{y} and \texttt{z} being the Euclidean coordinates in the front, left and up directions of the lidar itself, \texttt{t} $\in [0,10000000]$ being the nanoseconds since the start of the revolution, \texttt{reflectivity} $\in [0,255]$ being a measure of reflectivity of the surface that reflected the beam and \texttt{near\_ir} $\in [0,65535]$ being a measure of illumination.
The points of a lidar scan are ordered according to their beam of origin as well as their timestamp \texttt{t} within the scan. To fully preserve the structure of the scan, we included missing reflections through points containing all \texttt{NaN} values. 

\subsubsection{Ground Truth data format} \hfill\\
For the convenient evaluation of new algorithms, we provide the ground truth poses synchronized with the acquisition times of the lidar point clouds as a \ac{wsv} file \texttt{refsys/ground\_truth\_poses.txt}. The data format is identical to the KITTI data format, where every row contains the first three rows of a $4\times4$ homogenous transformation matrix specifying the orientation and translation of the vehicle w.r.t. the first pose, which is always set to the identity matrix. The basis for this data was acquired from the RTK/INS reference system (Section~\ref{sec:reference_system}) and postprocessed using loop closure constraints as described in Section~\ref{sec:ground_truth}. For time synchronization we utilized the timestamp acquisition time of the lidar scan and matched them to the closest measurement of the reference system. With a frequency of $\SI{200}{\hertz}$ for the reference system, this process results in a maximum absolute time difference of \SI[parse-numbers=false]{1/(200 \cdot 2)}{\second} $=$ \SI{2.5}{\milli\second}.

\section{EVALUATION}\label{sec:evaluation}
For the validation of our data and for easy comparison for future work, we provide the results of a multitude of publicly available localization methods on our data. These include the two lidar-only odometry methods
KISS-ICP~\citep{vizzo2023kiss} and MAD-ICP~\citep{ferrari2024mad}, the \ac{lio} approach RKO-LIO~\citep{malladi2025robust} as well as the map-assisted lidar-odometry approach OSM-ICP~\citep{kurda2025reducing}. For all methods we used the default parameter configuration as a baseline, modifying only the maximum and minimum ranges to $\SI{160}{\meter}$ and $\SI{3}{\meter}$ respectively, and enabling scan deskewing as a preprocessing step. All experiments were conducted on a laptop equipped with an Intel Core i7-1370P CPU and 32 GB of RAM. All data was stored on an external SSD. Under the chosen parameter settings, KISS-ICP achieved the highest throughput of $34.36$ FPS, followed by RKO-LIO and OSM-ICP with $17.37$ and $7.64$ FPS and lastly MAD-ICP with $6.44$ FPS. We measured the accuracy for all methods using three different metrics: the \ac{rpe} over sequences of the length of $\SI{100}{\meter}$ to evaluate the drifting behavior, the \ac{ape} for measuring the global consistency and lastly the frequently used KITTI metric~\citep{geiger2013vision} that, similarly to the \ac{rpe}, measures the drifting behavior of the system, but evaluates sequences of different lengths of up to $800$ meters. For the calculation of the \ac{rpe} and \ac{ape} we used the evo Python package\footnote{\url{https://github.com/MichaelGrupp/evo}}, and for the calculation of the KITTI metric we used the official KITTI development kit\footnote{\url{https://www.cvlibs.net/datasets/kitti/eval_odometry.php}}. Table~\ref{tab:results} shows the results for all methods and all individual sequences. 

The KITTI errors for all methods lie between $0.39$ and $1.30$ (excluding \texttt{Tunnel} and other outliers), placing their overall performance within a range comparable to results reported on other datasets. 
None of the approaches stand out as particularly problematic.
The two \ac{gnss}-denied scenarios, \texttt{ParkingGarage} and \texttt{UndergroundCarPark}, do not exhibit any notable increase in difficulty; their error metrics remain well within the expected variance of the dataset. Overall, none of the sequences stand out as particularly problematic except \texttt{Tunnel}. All methods report increased error values for both the KITTI and the \ac{rpe} metric on this trajectory with the highest increase experienced on \texttt{Tunnel1}. A close inspection of the sequence reveals all methods to fail inside the tunnel while accelerating in stop-and-go traffic. A high amount of dynamic objects moving as a coherent mass causes the methods, wrongfully register them, instead of the (static) background. This problem is resolved for all methods once the tunnel is exited and the amount of static geometry increases. Even tunnels with little to no traffic can pose significant challenges with MAD-ICP and KISS-ICP additionally failing in \texttt{Tunnel2} and KISS-ICP and OSM-ICP failing inside the tunnel on \texttt{HighwayTunnel3}. In Table~\ref{tab:results}, we marked all sequences/method combinations that exhibit this behavior in red.
\input{tables/results_4}

\section{CONCLUSION}\label{sec:conclusion}
In this work, we have presented Odyssey, a new dataset tailored towards lidar and lidar-inertial and map-assisted lidar odometry. Its large scope covers a majority of standard driving situations from urban to rural situations. 
Through the use of a modern \ac{rlg}-based \ac{ins} we are additionally capable of providing data from genuine \ac{gnss}-denied environments including tunnels and, unprecedented, indoor parking garages. 
While the latter showed no notable increase in difficulty for lidar and lidar-inertial odometry methods, tunnels continue to pose substantial challenges.
We provided multiple trajectories that caused state-of-the-art lidar and lidar-odometry methods to completely fail or at from decreased accuracy in these situations, revealing the necessity for the development of more robust approaches.
\section*{Acknowledgement}

This work was funded by the German Federal Ministry for Economic
Affairs and Climate Action (BMWK) within the research project “OKULAr” (Grant No. 19A22003C).

\bibliographystyle{SageH}
\bibliography{literature.bib}

\end{document}

%% file: tables/overview_trajectories.tex
\begin{table*}[t]
\centering

\caption{Summary statistics and short description for our 12 different trajectories. 
Length and Duration are averaged between individual repetitions and rounded while 
Total Sum corresponds to the sum of length and durations of the entire dataset.}
\label{tab:trajectories}
\begin{tabular}{r|cc|llc}
\thline
Name & Length / m & Duration / s & Environment & Characteristics & Loop Closure\\ \hline
\texttt{Beltway} & 1751 & 221 & urban & high speed, heavy traffic & \xmark \\
\texttt{CountryRoad} & 8764 & 700 & rural & open field & \checkmark \\
\texttt{ForestRoad} & 1636 & 346 & rural & rough terrain, vegetation & \xmark \\
\texttt{Highway} & 12079 &  652 & highway & high speed & \xmark  \\
\texttt{HighwayTunnel} & 19465 & 900 & highway & high speed, long tunnel & \xmark  \\
\texttt{InnerCity} & 1251 & 369 & urban & heavy traffic & \xmark  \\
\texttt{Marketplace} & 868 & 221 & sub-urban & rough terrain & \checkmark  \\
\texttt{ParkingGarage} & 607 & 254 & sub-urban & indoor, spiral ramp & \checkmark  \\
\texttt{Suburb} & 4632 & 410 & sub-urban & heavy traffic & \xmark  \\ 
\texttt{Theatre} & 2549 & 351 & urban & heavy traffic & \xmark  \\
\texttt{Tunnel} & 470 & 305 & urban & stop-and-go, tunnel & \xmark  \\
\texttt{UndergroundCarPark} & 295 & 161 & sub-urban & indoor & \checkmark  \\ \hline \hline
Total Sum & 163099 & 14670 & & \\ 
\thline
\end{tabular}
\end{table*}

%% file: tables/overview_datasets_and_ins.tex
\begin{table*}[t]
\centering
\caption{Overview of automotive datasets and their used gyroscopes. An \xmark~denotes a missing secondary \ac{imu}}
\label{tab:datasets}
\resizebox{0.9\textwidth}{!}{%
\begin{tabular}{r|cc|c|c}
\thline
Dataset & \multicolumn{2}{c|}{\pfirst~ Gyro} & \psecond~ Gyro & Environment \\
    & Type & Bias Instability [\SI{}{\degree/h}] & Bias Instability [\SI{}{\degree/h}] & \\ \hline
KITTI~\citep{geiger2013vision} & \acrshort{mems} & $2$ & \xmark & Urban \\
Complex Urban~\citep{jeong2019complex} & \acrshort{fog} & $0.05$ & $10$ & Urban \\
MulRan~\citep{kim2020mulran} & \acrshort{fog} & $0.05$ & $10$ & Urban \\
UrbanNav~\citep{hsu2023hongkong} & \acrshort{fog} & $1$ & $18$ & Urban \\
HeLIPR~\citep{jung2024helipr} & \acrshort{mems} & $0.45$ & $10$ & Urban \\
EU Long Term~\citep{yan2020eu} & \acrshort{mems} & $20$ & \xmark & Urban \\ 
Ours & \acrshort{rlg} & $0.001$ & $0.8$ & Diverse \\
\thline
\end{tabular}}
\end{table*}

%% file: tables/results_4.tex
\begin{table*}[t]
\centering

\caption{Results for a variety of different lidar, lidar-inertial and map assisted lidar odometry methods. We report the errors using the \acrfull{ape}, the \acrfull{rpe} over trajectory length of $\SI{100}{\meter}$ and the KITTI error metric. \textcolor{red}{Red} sequences indicate sequences with a significant deviation from the ground truth in individual situations, e.g., tunnels.}
\label{tab:results}
\resizebox{\textwidth}{!}{%
\begin{tabular}{l|lll|lll||lll||lll}
\toprule
Sequence & \multicolumn{3}{c|}{KISS-ICP} & \multicolumn{3}{c||}{MAD-ICP} & \multicolumn{3}{c||}{RKO-LIO} & \multicolumn{3}{c}{OSM-ICP} \\ 
 & \multicolumn{3}{c|}{\citep{vizzo2023kiss}} & \multicolumn{3}{c||}{\citep{ferrari2024mad}} & \multicolumn{3}{c||}{\citep{malladi2025robust}} & \multicolumn{3}{c}{\citep{kurda2025reducing}} \\ 
 & APE & RPE & KITTI & APE & RPE & KITTI & APE & RPE & KITTI & APE & RPE & KITTI \\
\midrule
Beltway1 & 10.08 & 1.13 & 1.05 & 11.89 & 1.16 & 1.08 & 11.59 & 1.19 & 1.04 & 8.27 & 0.82 & 0.68 \\
Beltway2 & 10.15 & 1.13 & 1.05 & 11.22 & 1.15 & 1.0 & 7.03 & 1.18 & 1.0 & 8.47 & 0.78 & 0.66 \\
Beltway3 & 11.74 & 1.14 & 1.12 & 11.51 & 1.19 & 1.1 & 6.96 & 1.19 & 1.05 & 8.49 & 0.81 & 0.69 \\
CountryRoad1 & 108.38 & 1.12 & 1.2 & 45.86 & 1.19 & 1.03 & 91.49 & 1.18 & 1.02 & 68.41 & 0.75 & 0.73 \\
CountryRoad2 & 129.82 & 1.11 & 1.27 & 56.46 & 1.2 & 1.07 & 110.53 & 1.19 & 1.04 & 79.45 & 0.75 & 0.76 \\
CountryRoad3 & 135.24 & 1.12 & 1.3 & 51.03 & 1.2 & 1.03 & 109.25 & 1.19 & 1.04 & 81.09 & 0.75 & 0.78 \\
ForestRoad1 & 8.76 & 0.76 & 1.18 & 6.21 & 0.84 & 0.76 & 8.09 & 0.76 & 0.88 & 6.48 & 0.61 & 0.84 \\
ForestRoad2 & 8.74 & 0.76 & 1.29 & 4.24 & 0.78 & 0.73 & 5.25 & 0.74 & 0.94 & 5.87 & 0.57 & 0.87 \\
ForestRoad3 & 9.63 & 0.83 & 1.28 & 5.24 & 0.92 & 0.79 & 12.43 & 0.84 & 0.95 & 6.22 & 0.69 & 0.88 \\
Highway1 & 818.98 & 0.7 & 0.87 & 313.52 & 0.8 & 0.69 & 518.07 & 0.79 & 0.61 & 611.09 & 1.81 & 1.26 \\
Highway2 & 793.18 & 0.71 & 0.86 & 342.51 & 0.81 & 0.7 & 487.34 & 0.82 & 0.62 & 433.7 & 0.59 & 0.48 \\
Highway3 & 734.19 & 0.7 & 0.8 & 326.22 & 0.79 & 0.69 & 268.7 & 0.76 & 0.62 & 428.61 & 0.6 & 0.5 \\
HighwayTunnel1 & 1319.13 & 0.66 & 1.04 & 666.93 & 0.75 & 0.68 & 656.39 & 0.75 & 0.74 & 601.89 & 0.62 & 0.58 \\
HighwayTunnel2 & 1136.33 & 0.68 & 0.93 & 609.4 & 0.77 & 0.68 & 590.45 & 0.77 & 0.71 & 542.41 & 0.67 & 0.59 \\
HighwayTunnel3 & \red{1400.91} & \red{3.42} & \red{2.57} & 700.29 & 0.76 & 0.68 & 700.68 & 0.76 & 0.76 & \red{799.04} & \red{5.35} & \red{2.93} \\
InnerCity1 & 5.96 & 0.74 & 0.69 & 3.26 & 0.74 & 0.72 & 2.9 & 0.81 & 0.7 & 5.31 & 0.59 & 0.51 \\
InnerCity2 & 3.47 & 0.93 & 0.73 & 2.96 & 0.92 & 0.77 & 62.56 & 1.0 & 0.81 & 2.77 & 0.73 & 0.52 \\
InnerCity3 & 4.8 & 0.85 & 0.72 & 3.11 & 0.83 & 0.69 & 12.6 & 0.88 & 0.64 & 2.48 & 0.62 & 0.46 \\
Marketplace1 & 2.46 & 0.91 & 0.85 & 1.97 & 0.93 & 0.88 & 4.83 & 0.95 & 0.86 & 2.17 & 0.62 & 0.65 \\
Marketplace2 & 1.57 & 0.99 & 0.87 & 2.03 & 1.01 & 0.96 & 1.29 & 1.02 & 0.87 & 1.24 & 0.65 & 0.54 \\
Marketplace3 & 1.89 & 1.13 & 0.99 & 1.88 & 1.15 & 1.03 & 1.58 & 1.17 & 0.98 & 1.28 & 0.7 & 0.59 \\
ParkingGarage1 & 0.58 & 0.94 & 0.66 & 0.59 & 0.89 & 0.62 & 1.26 & 0.89 & 0.63 & 0.6 & 0.56 & 0.39 \\
ParkingGarage2 & 0.47 & 1.13 & 0.78 & 0.49 & 0.89 & 0.62 & 0.5 & 0.94 & 0.66 & 0.39 & 0.63 & 0.43 \\
ParkingGarage3 & 0.33 & 0.97 & 0.66 & 0.53 & 0.9 & 0.61 & 0.43 & 0.91 & 0.62 & 0.23 & 0.59 & 0.4 \\
Suburb1 & 128.54 & 0.64 & 1.01 & 68.81 & 0.71 & 0.8 & 91.01 & 0.71 & 0.69 & 83.37 & 0.52 & 0.62 \\
Suburb2 & 122.58 & 0.68 & 1.01 & 73.71 & 0.74 & 0.87 & 131.81 & 0.75 & 0.73 & 78.8 & 0.54 & 0.63 \\
Suburb3 & 119.29 & 0.65 & 0.97 & 84.89 & 0.71 & 0.88 & 143.36 & 0.71 & 0.73 & 80.98 & 0.52 & 0.6 \\
Theater1 & 23.16 & 0.88 & 0.85 & 20.4 & 0.91 & 0.8 & 60.88 & 0.93 & 0.73 & 16.54 & 0.65 & 0.55 \\
Theater2 & 29.53 & 0.88 & 0.86 & 26.65 & 0.9 & 0.79 & 49.47 & 0.93 & 0.73 & 20.93 & 0.62 & 0.59 \\
Theater3 & 22.62 & 0.92 & 0.85 & 18.81 & 0.92 & 0.84 & 18.31 & 0.98 & 0.74 & 17.34 & 0.67 & 0.59 \\
Tunnel1 & \red{28.91} & \red{18.82} & \red{26.21} & \red{19.9} & \red{21.88} & \red{26.77} & \red{31.1} & \red{16.81} & \red{26.25} & \red{10.55} & \red{3.94} & \red{10.37} \\
Tunnel2 & \red{24.2} & \red{10.9} & \red{11.48} & \red{33.51} & \red{19.27} & \red{15.14} & 2.42 & 1.83 & 1.76 & 1.03 & 0.96 & 0.73 \\
Tunnel3 & 9.67 & 1.55 & 3.83 & 3.55 & 2.01 & 1.64 & 2.69 & 1.02 & 1.65 & 2.17 & 0.91 & 0.95 \\
UndergroundCarPark1 & 0.34 & 0.9 & 0.73 & 0.58 & 0.86 & 0.69 & 0.73 & 0.87 & 0.71 & 0.21 & 0.58 & 0.49 \\
UndergroundCarPark2 & 0.4 & 0.99 & 0.84 & 0.58 & 1.0 & 0.83 & 0.78 & 0.96 & 0.82 & 0.31 & 0.64 & 0.56 \\
UndergroundCarPark3 & 0.35 & 0.93 & 0.78 & 0.61 & 0.94 & 0.79 & 0.74 & 0.9 & 0.77 & 0.28 & 0.59 & 0.51 \\
\bottomrule
\end{tabular}}
\end{table*}